\documentclass{article}

\usepackage{microtype}
\usepackage{graphicx}
\usepackage{subfigure}
\usepackage{booktabs} 
\usepackage{enumitem}

\usepackage{hyperref}

\usepackage[font=small,labelfont=bf]{caption}
\usepackage[preprint]{icml2026}

\usepackage{amsmath}
\usepackage{amssymb}
\usepackage{mathtools}
\usepackage{amsthm}

\usepackage[capitalize,noabbrev]{cleveref}

\usepackage{xcolor}

\theoremstyle{plain}

\theoremstyle{definition}

\theoremstyle{remark}

\usepackage[textsize=tiny]{todonotes}

\icmltitlerunning{Learning a Latent Pulse Shape Interface for Photoinjector Laser Systems}

\begin{document}

\twocolumn[

\icmltitle{Learning a Latent Pulse Shape Interface for Photoinjector Laser Systems}



\icmlsetsymbol{equal}{*}

\begin{icmlauthorlist}
\icmlauthor{Alexander Klemps}{tuhh}
\icmlauthor{Denis Ilia}{desy}
\icmlauthor{Pradeep Kr. Banerjee}{tuhh}
\icmlauthor{Ye Chen}{desy}
\icmlauthor{Henrik Tünnermann}{desy}
\icmlauthor{Nihat Ay}{tuhh}
\end{icmlauthorlist}

\icmlaffiliation{tuhh}{Institute for Data Science Foundations, Hamburg University of Technology, Germany}
\icmlaffiliation{desy}{Deutsches Elektron-Synchrotron (DESY), Hamburg, Germany}

\icmlcorrespondingauthor{Alexander Klemps}{alexander.klemps@tuhh.de}

\icmlkeywords{Machine Learning, ICML}

\vskip 0.3in
]



\printAffiliationsAndNotice{}  

\begin{abstract}
Controlling the longitudinal laser pulse shape in photoinjectors of Free-Electron Lasers is a powerful lever for optimizing electron beam quality, but systematic exploration of the vast design space is limited by the cost of brute-force pulse propagation simulations. We present a generative modeling framework based on Wasserstein Autoencoders to learn a differentiable latent interface between pulse shaping and downstream beam dynamics. Our empirical findings show that the learned latent space is continuous and interpretable while maintaining high-fidelity reconstructions. Pulse families such as higher-order Gaussians trace coherent trajectories, while standardizing the temporal pulse lengths shows a latent organization correlated with pulse energy. Analysis via principal components and Gaussian Mixture Models reveals a well-behaved latent geometry, enabling smooth transitions between distinct pulse types via linear interpolation. The model generalizes from simulated data to real experimental pulse measurements, accurately reconstructing pulses and embedding them consistently into the learned manifold. Overall, the approach reduces reliance on expensive pulse-propagation simulations and facilitates downstream beam dynamics simulation and analysis.

\end{abstract}

\section{Introduction}
\label{sec:introduction}

Modern free-electron lasers are inherently multi-stage systems, in which tightly coupled subsystems jointly determine beam quality. In particular, the temporal shape of the photocathode laser pulse responsible for the electron generation plays a critical role in defining the phase space of the emitted electron bunch, thereby influencing downstream beam dynamics, emittance, and stability \cite{stephan2016high, lemonsdispersion}. While this coupling offers powerful control opportunities, it also introduces a fundamental bottleneck: the space of physically attainable pulse shapes at the photocathode is high-dimensional and nonlinear, complicating targeted exploration. As a result, exploration typically relies on brute-force sampling strategies that become prohibitively expensive when combined with beam dynamics simulations, particularly in the absence of differentiable pulse propagation models.

Machine learning has emerged as a promising tool to alleviate these challenges, but existing approaches largely address individual subsystems in isolation. On the one hand, surrogate models and data-driven predictors have been developed to approximate beam dynamics and enable fast tuning of accelerator parameters. On the other hand, machine learning has been applied to laser pulse shaping and reconstruction, learning mappings between optical filter parameters and on-cathode pulse profiles. However, these approaches generally rely on fixed or hand-crafted representations at the interface between laser and accelerator domains, limiting their ability to support beam analysis and end-to-end optimization across system boundaries with required physical fidelity.

A central challenge therefore remains unresolved: how to construct a compact yet expressive and physically meaningful representation of upstream laser pulse shapes that can serve as an effective interface for downstream beam analysis and learning tasks, without requiring repeated access to laser propagation simulations and connected underlying physics. Addressing this challenge requires not only accurate reconstruction capabilities but also latent representations that are smooth, interpretable, and geometrically well-behaved, so that transitions between different families of pulse shapes and sampling from these remain physically plausible.

In this work, we propose a generative representation learning framework that addresses this interface problem by learning low-dimensional latent representations of temporal laser pulse shapes from simulated data. Using Wasserstein Autoencoders (WAEs) \cite{TolBouGelSch18}, we obtain latent manifolds that preserve the continuous structure of physically relevant and realizable pulse families while enabling high-fidelity reconstruction. The learned latent space exhibits clear semantic organization, with coherent trajectories corresponding to parameterized pulse families and dominant latent directions correlated with physically meaningful quantities such as pulse energy and transverse emittance.

Next to reconstruction quality, we analyze the geometry of the learned latent space and demonstrate that it supports physically plausible transitions w.r.t. the 2-Wasserstein distance between distinct pulse types via latent interpolation. Importantly, the model generalizes beyond the simulated domain to experimental laser pulse measurements, embedding real data consistently into the learned manifold. This establishes the latent space as a robust, data-driven interface between laser shaping and beam dynamics.

\textbf{Main contributions.} Our contributions are threefold:
\begin{itemize}[leftmargin=*]
\item We introduce a generative modeling approach that learns a low-dimensional, differentiable latent interface between longitudinal laser pulse shaping and beam dynamics, enabling efficient exploration of pulse shape variations,
\item We demonstrate that WAEs yield smooth and interpretable latent geometries in a Wasserstein sense in which distinct pulse families are organized along coherent trajectories, allowing artifact-free interpolation and analysis of pulse morphologies, and
\item We show that the learned representations generalize to experimental pulse measurements and support large-scale beam dynamics simulations, revealing latent directions that strongly correlate with transverse emittance in the near-cathode regime.

\end{itemize}

\section{Related Work}
\label{sec:related_work}

Machine learning is increasingly used to address core challenges in accelerator science, particularly in modeling and controlling electron beam quality. Existing research has largely focused on two distinct areas: (i) surrogate modeling of accelerator beam dynamics and (ii) inverse modeling or control of laser pulse shaping based on measurements.

In surrogate modeling, the goal is to replace computationally expensive calculations in simulations with fast, data-driven approximations. Numerous studies in beam dynamics such as \cite{xu2022surrogate} have trained neural networks to predict beam properties—such as emittance, bunch length, and transverse profiles—based on machine settings and, in some cases, measured inputs like laser profiles. In the domain of laser shaping, \cite{hirschman2025lstm} developed an LSTM-based surrogate for nonlinear optical propagation and integrated it into a start-to-end simulation pipeline for LCLS-II, replacing computation heavy fiber propagation simulations with a fast running model of high fidelity. Other approaches leverage the implementation of differentiable simulation codes as in \cite{kaiser2022learning, kaiser27bridging} and \cite{iliadifferentiable}. Such models serve as digital twins for online photon laser and electron beam tuning and uncertainty quantification.

\begin{figure}
    \centering
    \includegraphics[width=\linewidth]{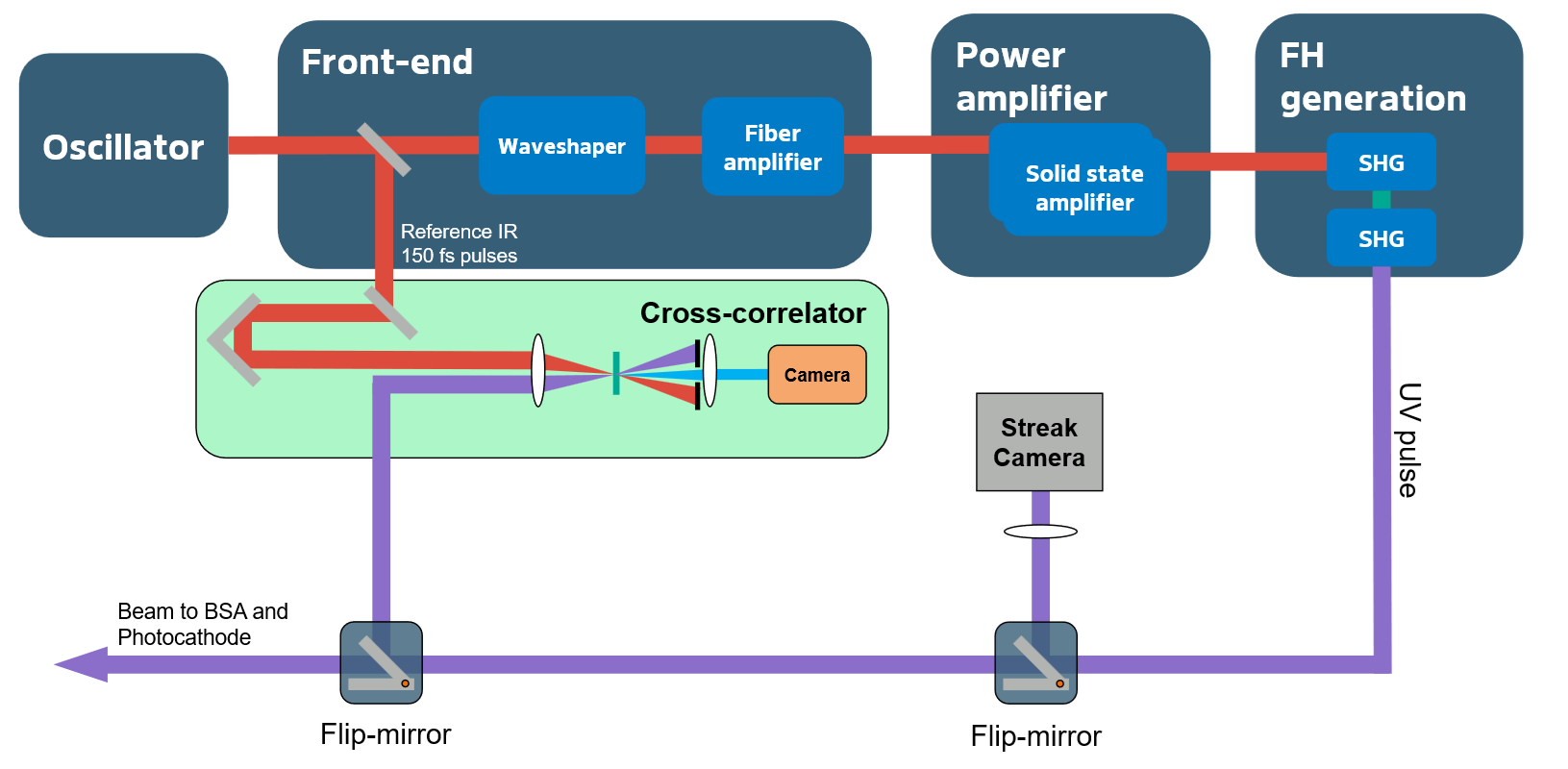}
    \caption{Layout of the modeled photoinjector laser system. Initially in the front-end created pulses undergo nonlinear transformations during propagation to the photoinjector system.}
    \label{fig:nepal}
\end{figure}

Complementary work addresses inverse problems in laser shaping: controlling or reconstructing laser pulse shapes to achieve desired outcomes. In ~\cite{pollard2022temporal} and ~\citep{xu2021slm} neural network–based control of spatial light modulators and pulse shapers for tailoring temporal or spatiotemporal pulse profiles in photoinjector systems have been demonstrated. And the idea of controlling pulse shape characteristics via low dimensional representations has recently been expressed in \cite{gutierrezdifficf}. These approaches aim to bypass slow physics-based propagation models by learning effective mappings from on-cathode pulse forms to shaping parameters, accounting for nonlinear effects such as harmonic conversion and fiber propagation.

A recent review of generative modeling in accelerator applications~\cite{annurev:/content/journals/10.1146/annurev-nucl-121423-100719} highlights the growing use of Autoencoders (AEs), Variational Autoencoders (VAEs), and Generative Adversarial Networks (GANs) for learning low-dimensional representations of beam distributions, phase space, or machine configurations. These methods enable efficient sampling, compression, and simulation-speedup across a variety of facilities and beamlines. Notably, while VAEs and GANs have been used to model predict phase spaces~\cite{edelen2018vae, zhu2021high} and transfer functions~\cite{scheinker2021gan}, WAEs have not yet received significant attention in this domain, despite their theoretical strengths in distribution matching and manifold regularity \cite{rubenstein2018latentspacewassersteinautoencoders}.

Recently, the interface between the two domains of laser shaping and beam dynamics has gained increasing attention. \cite{bhpv-bcqk} present a GAN-based approach to predict resulting space charge forces from Gaussian transverse laser shape input. \cite{gupta2021lcls} used a convolutional encoder–decoder architecture to model the LCLS-II injector, incorporating full laser distribution images on the cathode as inputs to predict downstream beam observables. While such pipelines are highly promising, they typically only cover one or two parametric families of pulse shapes at the photocathode. By learning latent representations of diverse families of laser pulse shapes, our work targets this emerging niche and opens a promising direction for beam dynamics analysis and modeling via a well-behaved latent space.

\section{Pulse Shape Data}
\label{sec:data}

In ultrafast optics \cite{Weiner2009}, a laser pulse is typically described in the spectral domain by its complex electric field 
\begin{align}
    E(\omega)=A(\omega)e^{i\phi(\omega)}\label{eq:pulse}
\end{align}
where $A(\omega)$ is the spectral envelope and $\phi(\omega)$ the spectral phase. 
We expanded the spectral phase in a polynomial around the central frequency of $\omega_0$ such that
\[
\phi(\omega) = \frac{1}{2}\varphi_2(\omega-\omega_0)^2 + 
    \frac{1}{6}\varphi_3(\omega-\omega_0)^3 + 
    \frac{1}{24}\varphi_4(\omega-\omega_0)^4
\]
capturing second-, third- and fourth-order dispersions $\varphi_2,\,\varphi_3$ and $\varphi_4$.
To explore a broad family of physically relevant pulse shapes, we simulated input pulses at a waveshaping device within the front-end of the photoinjector laser system shown in Figure \ref{fig:nepal} by sampling over relevant shaping parameter ranges. Throughout all simulations $\omega_0$ corresponded to a central wavelength of $\lambda_0=1030\,\text{nm}$. The envelopes are chosen to be either secant (S), parabolic (P), flattop (F), triangular (T) or Gaussian (G) shaped, with triangular and Gaussian shapes additionally 
sampled at varying orders $p_T$ and $p_G$. An overview on the distributions, ranges and units of the involved parameters is given by Table \ref{tab:parameters}. \par

This sampling procedure yielded a rich set of complex spectra representative of realistic shaping scenarios at the laser front-end. We simulated their propagation through the laser system fiber front end using \textit{RP Fiber Power} \cite{RPFiberPower},  a commercial solver that couples the nonlinear Schrödinger equation with rate equations. While subsequent frequency conversion stages modify these profiles, the simulations provide a representative set of temporal shapes for beam dynamics studies. These simulations account for fiber dispersion, nonlinearities, and absorption, producing physically accurate temporal profiles as they would appear at the photocathode. \par

The obtained simulation outcomes have been condensed into a dataset containing a total of $10\,000$ pairs of complex spectra of input and corresponding propagated pulses, each represented on a spectral grid of size $2^{13}=8192$ with a resolution of $\Delta\omega\approx 1.041\,\text{GHz}$. The data is available upon request.

\begin{table}[t]
    \caption{Pulse parameters and sample ranges.}
    \centering
    \begin{tabular}{c|c|c}
    Parameter & Distribution & Unit \\ \hline
    $A(\omega)$ & $\mathcal{U}(\{S,P,F,T,G\})$ & -\\
    $p_T$ & $\mathcal{U}(\{1,2,4\})$ & - \\
    $p_G$ & $\mathcal{U}(\{1,2,3,4,5,10\})$ & - \\
    $\sigma_t$ & $\mathcal{U}([2,\,40])$ & ps \\
    $\varphi_2$ & $\mathcal{N}(0, \frac{100}{3\sigma_t^2})$ & $\text{s}^2$ \\
    $\varphi_3$ & $\mathcal{N}(0, \frac{100}{3\sigma_t^3})$ & $\text{s}^3$ \\
    $\varphi_4$ & $\mathcal{N}(0, \frac{400}{3\sigma_t^4})$ & $\text{s}^4$ \\    
    \end{tabular}
    \label{tab:parameters}
\end{table}

\subsection{Data Preprocessing}

\begin{figure}
    \centering
    \includegraphics[width=0.8\linewidth]{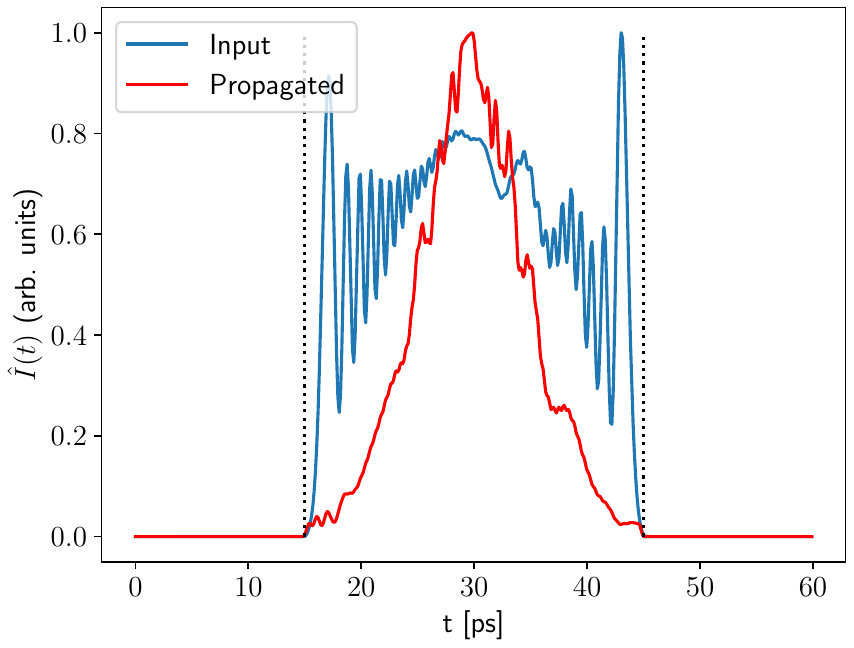}
    \caption{Representative simulated pulse pair before and after propagation through the fiber system. The input pulse (blue) and the propagated pulse (red) are shown as normalized intensity profiles after centroid and support alignment. Propagation leads to temporal reshaping due to dispersion and nonlinear effects, highlighting the variety of pulse morphologies within the dataset.}
    \label{fig:pulse_preprocessed}
\end{figure}

In practice, only the temporal intensity profile
\[
I(t) = |E(t)|^2 = \Big|\mathcal{F}^{-1}\{E(\omega)\}\Big|^2
\]
is accessible via measurements. For the purposes of this work, each pulse is therefore identified with its time-domain intensity profile. In order to remove trivial sources of variability and to standardize the input for learning, we apply the following preprocessing steps:
\begin{itemize}[leftmargin=0pt]
    \item[] \textbf{Peak normalization.} Intensities are normalized by their maximum value, $\hat{I}(t) := \frac{I(t)}{\max_t\,I(t)}$. This preserves differences in integrated pulse energy while removing trivial amplitude scaling.
    \item[] \textbf{Center-of-mass alignment.} Profiles are shifted such that the temporal centroids coincide with the center of the time grid, removing arbitrary timing offsets.
    \item[] \textbf{Support standardization.} The support lengths of all profiles are unified via polynomial interpolation to have equal length on the time grid, corresponding to a temporal pulse length of $30\,\text{ps}$.
    \item[] \textbf{Cropping.} Each profile is cropped around its centroid and represented on a time grid of size $512$, chosen as a trade-off between temporal resolution and computational efficiency.
\end{itemize}
An example of a preprocessed pair of pulses is shown in Figure \ref{fig:pulse_preprocessed}. Since our goal is to learn generalizable representations and characteristics of pulse shapes, for subsequent analysis we treat all pulses equally, regardless of whether they are recorded before or after fiber propagation.

\section{Methodology}
\label{sec:methods}

\subsection{Model Architecture}

Encoder–decoder models such as Autoencoders or Variational Autoencoders by \cite{kingma2014vae} and extensions such as $\beta$-VAES \cite{higgins2017beta} are widely used to learn low-dimensional representations of complex physical data. However, in our physical setting latent variables must preserve geometric continuity such as smooth transitions between pulse shapes. In those regards, per-sample regularization terms in VAEs can lead to over-regularized latent spaces and information loss.

\begin{figure*}
    \centering
    \includegraphics[width=\linewidth]{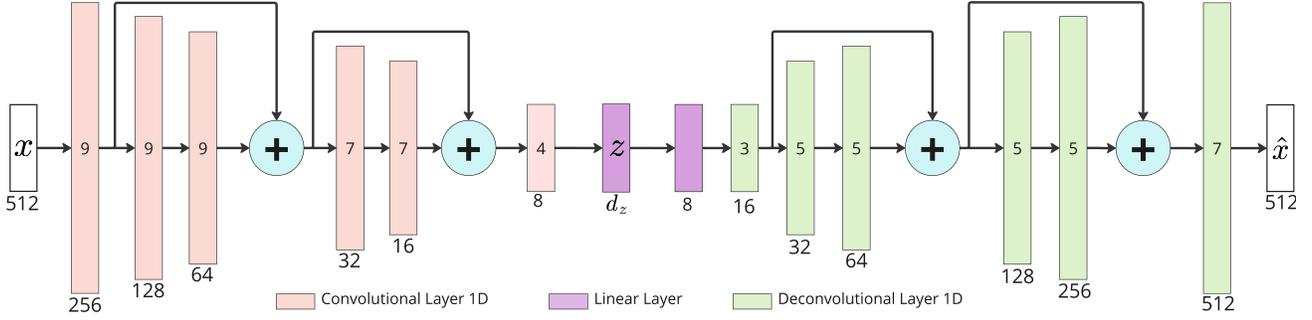}
    \caption{Structure of the convolutional encoder and decoder forming the Wasserstein Autoencoder. Each block consists of one-dimensional convolutions followed by batch normalization and Leaky-ReLU activations. Residual skip connections are included to stabilize training and preserve fine temporal details. Layer output sizes are shown below each block, and kernel sizes are indicated by numbers inside the convolutional layers. The decoder’s final output layer uses a hyperbolic tangent activation to constrain the reconstruction to the normalized signal range.}
    \label{fig:wae}
\end{figure*}

In this work, we therefore employ the deterministic variant of the WAE as introduced by \cite{TolBouGelSch18}. Both the encoder $E_\phi$ and decoder $G_\theta$ are deterministic, mapping pulse profiles $x\in\mathbb{R}^{512}$ distributed according to $X\sim p_X$ to
latent codes $z:=E_\phi(x)\in\mathbb{R}^{d_z}$ and reconstructions $\hat{x}:=G_\theta(z)$, where $d_z$ denotes the dimension of the latent space. Due to the data being one-dimensional signal vectors, encoder and decoder are implemented as Convolutional Neural Networks structured as shown in Figure \ref{fig:wae}.

\subsection{WAE Learning Objective}

Unlike VAEs, no stochastic sampling or reparameterization is performed within the encoder. Instead, stochasticity enters only through minibatch sampling and through draws from the prior distribution $p_Z=\mathcal{N}(0, I_{d_z})$ when matching with the aggregated posterior
\[
q_Z(z)=\int q_\phi(z|x)p_X(x)\,dx
\]
via some divergence \textit{D}. The overall WAE objective writes then as
\[
\mathcal{L}_{\text{WAE}} = 
    \mathbb{E}_{x\sim p_X}[c(x, G_\theta(E_\phi(x)))] + \lambda D(q_Z,p_Z)
\]
with transport cost function $c(x,\hat{x})$ and some weight parameter $\lambda>0$. In our deterministic setting and consistent with \cite{TolBouGelSch18}, we chose \textit{c} to be the MSE loss and \textit{D} to be the \textit{maximum mean discrepancy} (MMD), with an estimator given as
\begin{align}
\mathrm{MMD}^2(q_Z, p_Z)
= &
~\mathbb{E}_{z,z'\sim q_Z}\big[k(z,z')\big]
+\mathbb{E}_{\tilde{z},\tilde{z}'\sim p_Z}\big[k(\tilde{z},\tilde{z}')\big]\notag\\
& -2\,\mathbb{E}_{z\sim q_Z,\tilde{z}\sim p_Z}\big[k(z,\tilde{z})\big],\notag
\end{align}
where \(k(\cdot,\cdot)\) denotes a characteristic, positive definite kernel. In this work, we employ the \textit{inverse multiquadratic} (IMQ) kernel
\begin{equation}
\label{eq:imq_kernel}
k_{\mathrm{IMQ}}(z, z')
=
\frac{C}{C + \lVert z - z'\rVert_2^2},\notag
\end{equation}
with a scale parameter $C>0$ (averaged over several scales in practice), since it has been shown to provide a strong and smooth alignment between the aggregated posterior and the isotropic Gaussian prior $p_Z$.

\subsection{Sampling Electron Emission Times}
\label{subsec:linear_transform_sampling}

Aiming for interfacing with downstream beam dynamics simulations via the learned latent representations and reconstructions, we interpret the normalized longitudinal laser pulse \(I(t)\) as a probability density function over emission time. 
Specifically, we define
\[
p(t) := \frac{I(t)}{\int_{t_{\min}}^{t_{\max}} I(s)\,ds},
\qquad T \sim p(t),
\]
so that \(T\) is a random variable representing the electron emission time. 
The corresponding cumulative distribution function
\[
F(t) = \int_{t_{\min}}^{t} p(s)\,ds
\]
is monotonic and therefore admits a generalized inverse
\begin{align}
    F^{-1}(u) := \inf\{t \mid F(t) \ge u\}. \label{eq:gen_inverse}
\end{align}
This enables the sampling of electron emission times consistent from a laser pulse shape via Inverse Transform Sampling \cite{DEVROYE200683}. Given a distribution function $F$ and its inverse $F^{-1}$, a random variable \textit{T} following the distribution imposed by a laser pulse shape can be generated as \(T = F^{-1}(U)\), where \textit{U} is a random variable \(U \sim \mathcal{U}([0,1])\).
This construction provides a natural probabilistic interpretation of laser pulses and enables sampling of emission times directly from modeled pulse profiles. Examples of sampled electron emission times histograms from pulse shapes can be found in Appendix \ref{appendix:sampling}.

\section{Results}

\subsection{Training and Evaluation}

We implemented and trained the WAE using the outlined network architecture and objective in PyTorch, employing the framework \textit{Pythae} ~\cite{chadebec2022pythae}. The total of \(20\,000\) temporal intensity profiles were split into disjoint training and test sets in an \(80{:}20\) ratio. Training was performed for \(150\) epochs using the Adam optimizer with a learning rate of \(10^{-3}\) and a batch size of \(64\). In our experiments, a latent dimension of \(d_z = 32\) provided a good trade-off between compression and reconstruction accuracy. 
The regularization weight for the MMD term was set to \(\lambda = 0.1\), and the divergence was computed with an inverse multiquadratic kernel averaged over multiple kernel scales. The model comprised approximately \(2.4\times10^6\) trainable parameters across encoder and decoder, balancing expressive capacity and computational efficiency.  

\begin{figure*}
    \centering
    \includegraphics[width=\linewidth]{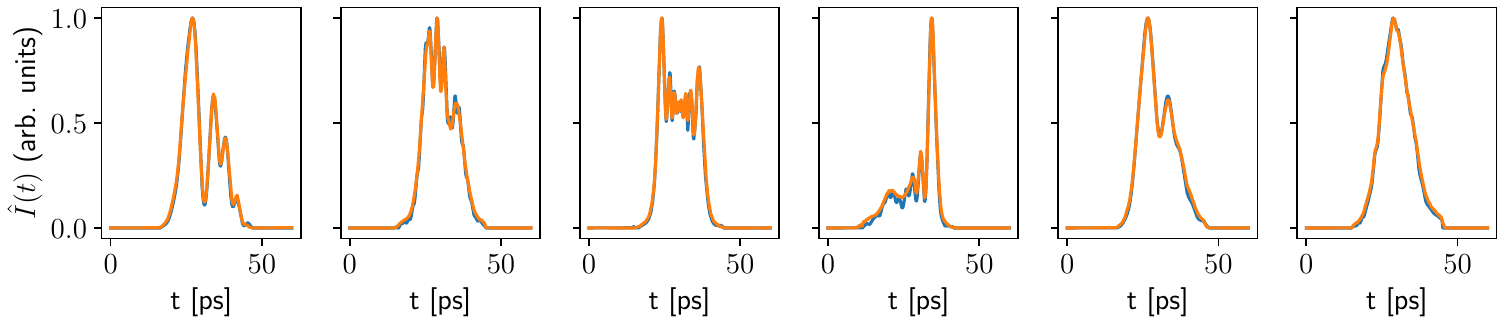}
    \caption{Comparison between input pulse profiles from the test dataset (blue) and their corresponding reconstructions (orange) obtained with the trained Wasserstein Autoencoder.
The reconstructions capture not only the overall pulse envelope but also higher-frequency oscillatory components, demonstrating the model’s ability to preserve fine temporal structure.}
    \label{fig:recon}
\end{figure*}

Training and validation losses converged smoothly without signs of overfitting. 
Reconstructions on the held-out test set reproduce both smooth and structured features of the input pulses with high accuracy, as shown in Figure~\ref{fig:recon}.

\subsection{Latent Space Structure}

\begin{figure}[!h]
    \centering
    \includegraphics[width=\linewidth]{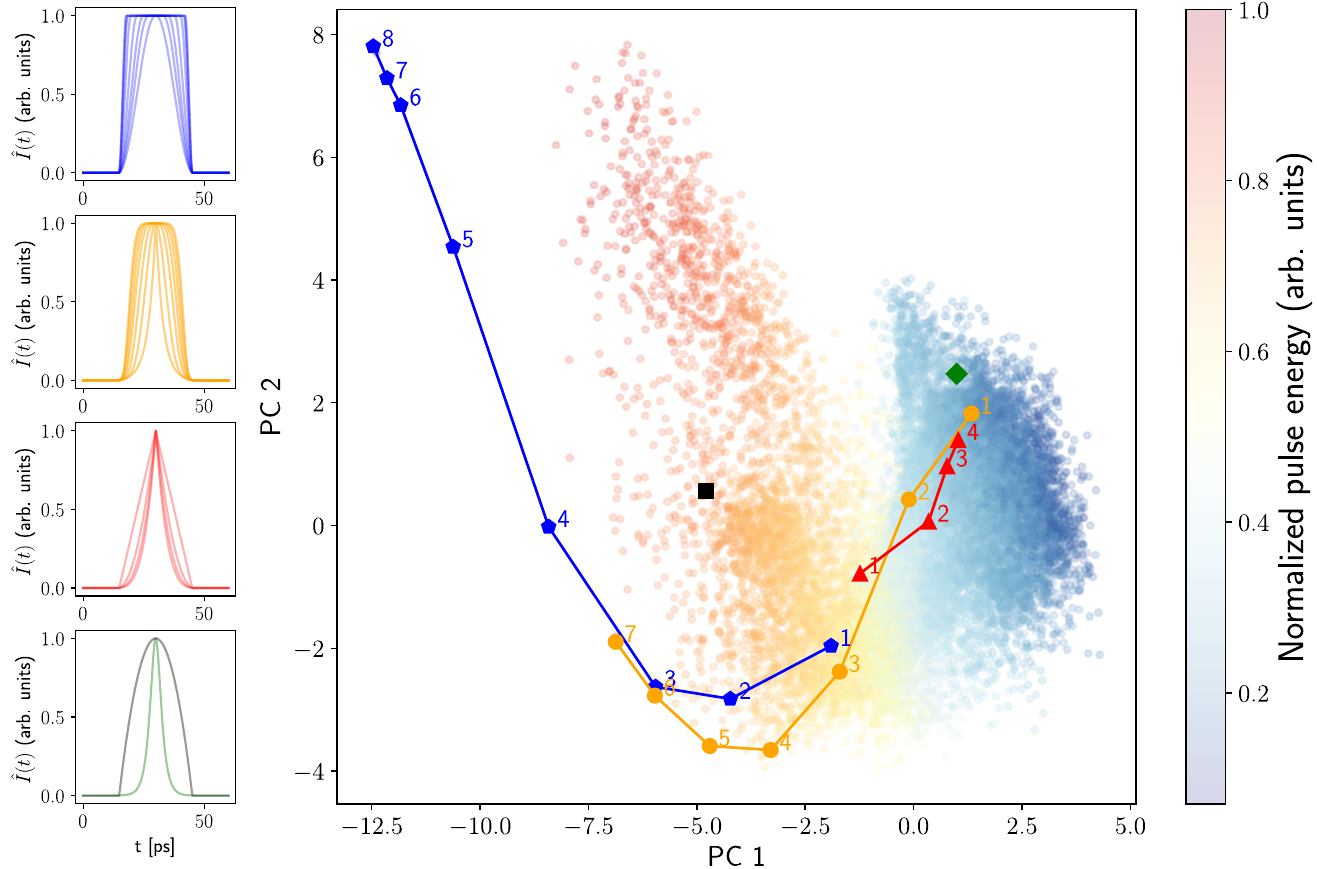}
    \caption{Visualization of the learned WAE latent space projected onto the first two principal components.
Each point represents an encoded pulse, color-coded by normalized pulse energy.
The overlaid trajectories correspond to parameterized pulse families shown on the left, whose latent embeddings evolve smoothly along coherent paths across the manifold.}
    \label{fig:latent_1}
\end{figure}

To study the organization of certain pulse shapes in the latent space, we analyzed how the WAE arranges distinct families of laser pulse profiles such as Gaussian, super-Gaussian, triangular, and flattop distributions. When projecting the latent codes onto the first two principal components, these families form coherent trajectories, as shown in Figure \ref{fig:latent_1}. In particular, Gaussian pulses of increasing order approach the region occupied by flattop distributions, reflecting the mathematical limit of super-Gaussian functions as their order parameter tends to infinity. Similarly, triangular and parabolic pulses populate nearby regions, indicating that the learned latent space captures a meaningful notion of pulse shape similarity.

Beyond these qualitative observations, quantitative analysis of the latent variables reveals further physically meaningful structure. One of the principal components exhibits a strong linear correlation with the total pulse energy, suggesting that the WAE encodes this physically relevant quantity as a dominant mode of variation. This correlation arises from the preprocessing, where all pulses share equal temporal support and amplitude scaling, allowing energy to emerge as the primary discriminating feature. Fitting a Gaussian Mixture Model (GMM) to the latent codes further highlights the structured organization of the space. Since we expect at least the five families of envelope shapes as stated in Table \ref{tab:parameters} to be present in the dataset and also to account for shapes transformed by nonlinear effects due to propagation, we decided to fit a GMM with six components to the latent space. Investigating the means of these components shows that the mixture components correspond to clusters of pulses of distinct shapes and energy content. Furthermore, measuring the pairwise $W_2$ distances between the components yields the flattop component as the one with the largest distance to all other ones, where the distance to the Gaussian component is the smallest. Additional details on the fit GMM can be found in Appendix \ref{appendix:gmm}.


\begin{figure*}[!h]
    \centering
    \includegraphics[width=\linewidth]{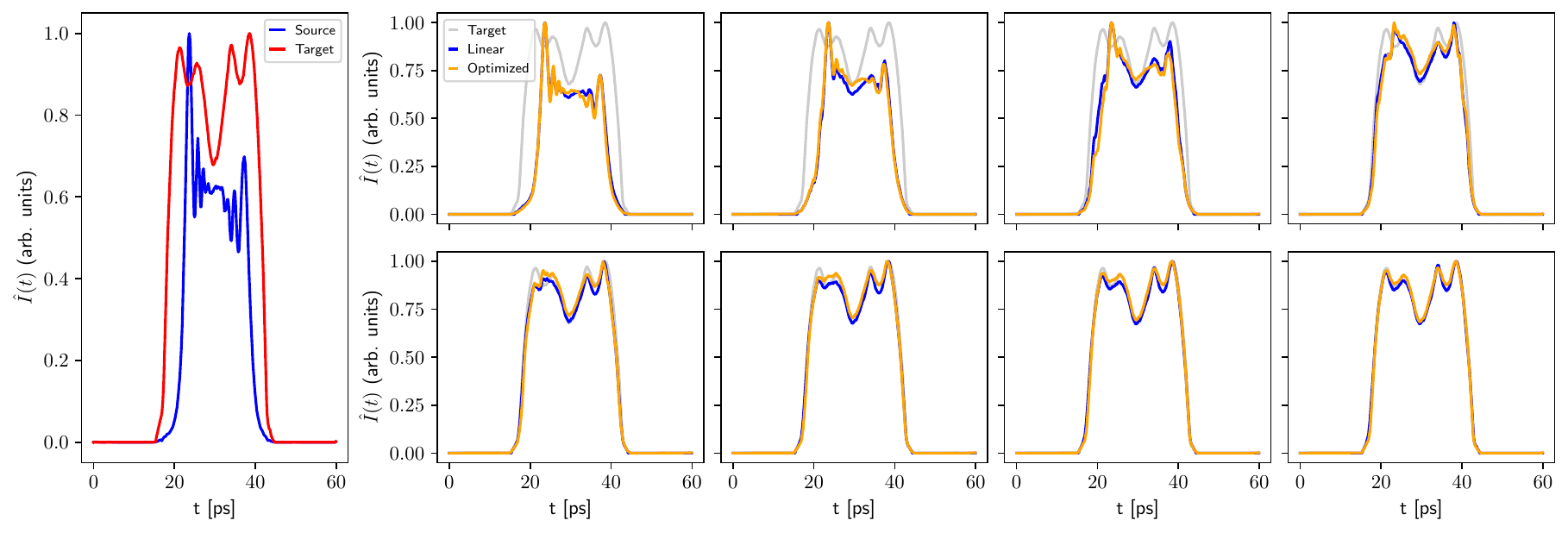}
    \caption{Latent interpolation between two pulse profiles. Left: source and target pulses with a $W_2$ distance of $1.080$. Right: sequence of intermediate decoded pulses from latent interpolations via a discrete linear interpolation path $\Gamma_0$ (blue) and a $W_2$-optimized path $\Gamma^*$ (orange) using $N=10$ waypoints. Both yield smooth transitions w.r.t. the $W_2$ distance, where the optimized interpolation path attains a near-optimal length of $L(\Gamma^*)=1.084$ compared to slightly higher $L(\Gamma_0)=1.195$.}
    \label{fig:latent_interpolation}
\end{figure*}

\subsection{Linear and Geometry-Aware Interpolation}
\label{sec:geometry_aware_interpolation}

Linear interpolation in latent space is a standard tool for probing the structure of generative models. 
However, straight lines in the Euclidean latent space $(\mathbb R^{d_z}, \|\cdot\|_2)$ need not necessarily correspond to semantically smooth or physically meaningful transitions in the data space of laser pulse profiles. It is therefore vital to respect the latent manifold's geometry when interpolating. Within prior work \cite{a3a0e62556004f0987c8551e3187adf8}, geodesic paths induced by decoders of under Euclidean or approximate Riemannian metrics are explored, and \cite{pmlr-v130-arvanitidis21a} demonstrates that optimized latent trajectories possibly better reflect semantic variation than linear interpolation alone. More recent research on geodesic calculus in latent spaces \cite{hartwig2025geodesiccalculuslatentspaces} presents a principled framework for computing and visualizing such discrete latent paths. We adopt a latent geodesic optimization paradigm with path length measured in the 2-Wasserstein distance, which provides a natural notion of intensity redistribution and pulse morphing. We then compare the resulting optimized interpolations to linear interpolations in latent space.

For each latent code \( z \in \mathbb{R}^{d_z} \), the decoded pulse profile \( G_\theta(z) \) corresponds by area-normalization to a probability density \( p_z(t) \) and the associated probability measure \( \mu_z \). By construction, \( \mu_z \) has compact support
\(
\operatorname{supp}(\mu_z) \subseteq [t_{\min}, t_{\max}],
\)
and therefore a finite second moment. Consequently, the 2-Wasserstein distance is well defined, and interpolation can be performed in the metric space \( (\mathcal P_2(\mathbb R), W_2) \), where \( \mathcal P_2(\mathbb R) \) denotes the set of probability measures on \( \mathbb R \) with finite second moment. As shown in \cite{7974883}, this space admits the structure of a formal infinite-dimensional Riemannian manifold, enabling geodesic calculus.

For any two measures \( \mu_{z_i}, \mu_{z_j} \in \mathcal P_2(\mathbb R) \), the 2-Wasserstein distance in one dimension has the closed-form expression
\[
W_2(\mu_{z_i}, \mu_{z_j})
= \left( \int_0^1 \bigl( F_{z_i}^{-1}(u) - F_{z_j}^{-1}(u) \bigr)^2 \, du \right)^{1/2},
\]
where \( F_z \) denotes the cumulative distribution function of \( p_z \), and \( F_z^{-1} \) its generalized inverse as in Eq.~\ref{eq:gen_inverse}.

To approximate a geodesic between two latent codes \( z' \) and \( z'' \), we initialize a path
\(
\Gamma = \{ z_1, \dots, z_N \}
\)
by linear interpolation, with fixed endpoints \( z_1 = z' \) and \( z_N = z'' \). We define the discrete path length
\[
L(\Gamma)
= \sum_{i=1}^{N-1} W_2\bigl( \mu_{z_i}, \mu_{z_{i+1}} \bigr),
\]
and minimize \( L(\Gamma) \) with respect to the intermediate points \( z_2, \dots, z_{N-1} \) using the Adam optimizer, while keeping the endpoints fixed. Gradients propagate through both the decoder and the empirical Wasserstein computation, yielding an optimized path \( \Gamma^\ast \) whose decoded pulse profiles approximate a geodesic in \( (\mathcal P_2(\mathbb R), W_2) \). An example for latent interpolation is provided by Figure \ref{fig:latent_interpolation}.

Given the computational optimization costs, we are interested in quantifying the advantage of optimized paths over linear interpolation. For each pair of latent endpoints \( (z', z'') \) and a connecting discrete path $\Gamma$, we quantify path optimality by the optimality ratio
$\rho(\Gamma) := L(\Gamma)/W_2(\mu_{z'}, \mu_{z''}) \ge 1$. Across 500 randomly sampled endpoint pairs, linear interpolation paths \( \Gamma_0 \) yield an average optimality ratio of $\rho(\Gamma_0) = 1.1442 \pm 0.1806$, whereas optimized paths \( \Gamma^\ast \) achieve ratios close to unity $\rho(\Gamma^\ast) = 1.0082 \pm 0.0127$. Although the optimization consistently reduces the accumulated path length, the quantitative improvement over linear interpolation is small relative to the additional computational effort. The narrow gap between \( \rho(\Gamma_0) \) and \( \rho(\Gamma^\ast) \) suggests that linear interpolation in the WAE latent space provides a sufficiently accurate and computationally efficient approximation. This near-optimality of linear paths indicates that the learned latent space is already well-regularized with respect to the 2-Wasserstein distance between pulses in data space.

\subsection{Generalization to Experimental Data}
\begin{figure*}
    \centering
    \includegraphics[width=\linewidth]{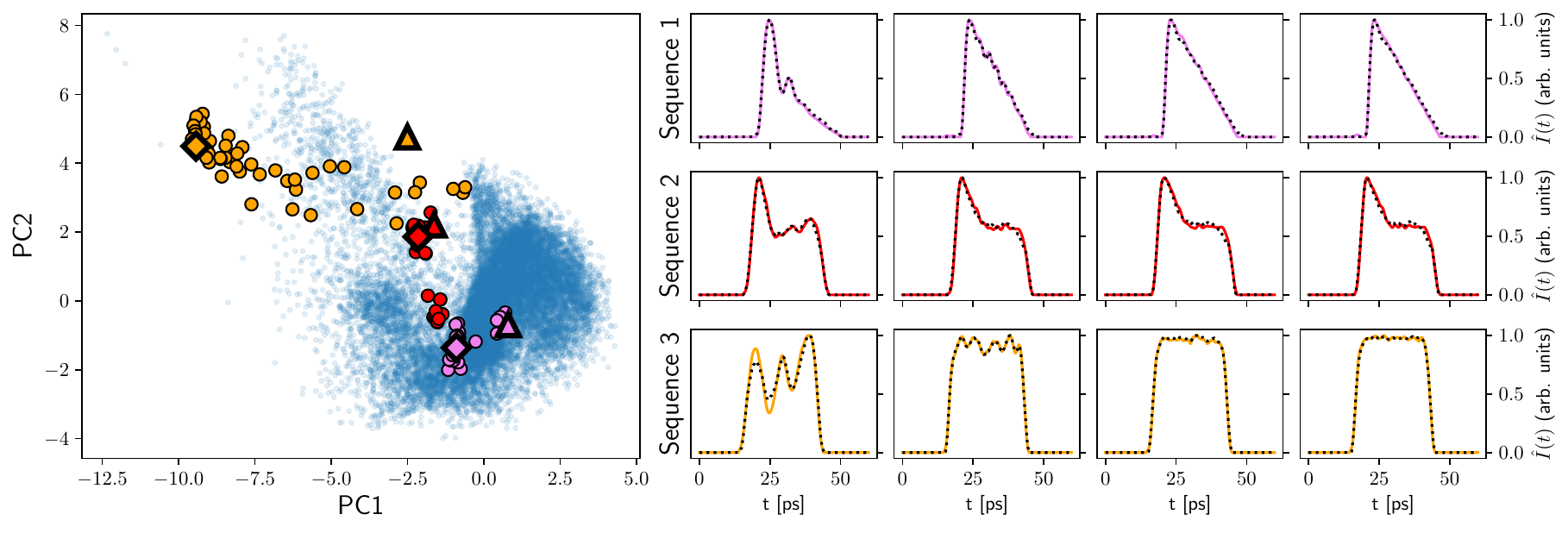}
    \caption{Comparison between experimentally measured and preprocessed pulse profiles (colored solid) and corresponding reconstructions (black dotted) produced by the trained Wasserstein Autoencoder. Each row to the right resembles representative samples from recorded sequences of pulses measured at the photoinjector laser system, in which optical filter parameters were varied through a gradient-based optimization to morph the initial profile (triangle/first column) into a desired target shape (square/last column). Pulses in the first two sequences are infrared ones, whereas the third one contains UV pulses. The reconstructions closely match the measured data, demonstrating high accuracy and generalization to real experimental conditions.
}
    \label{fig:exp}
\end{figure*}

To evaluate the model’s ability to generalize beyond the simulated domain, we applied the trained WAE to three sequences with a total of 109 experimentally measured infrared and ultraviolet pulse profiles from the 
photoinjector laser system. Figure \ref{fig:exp} compares measured pulses and their reconstructions.
While the infrared sequences lie within the high-density region of the latent manifold defined by the simulated training data, the ultraviolet pulses are mapped to sparsely populated regions at the periphery of the latent space.
Despite this extrapolation, the WAE reproduces the temporal structures of the experimental pulses with high fidelity, including fine-scale oscillations.
This indicates that the learned latent representation captures general physical relationships between pulse shapes rather than overfitting to the simulated distribution, supporting its use as a robust interface for experimental regimes not explicitly covered during training.

\begin{table}[t]
\caption{Comparison of $\beta$-VAEs for different $\beta$ and WAE on simulated test and experimental datasets. $\uparrow$ denotes that higher values indicate better performance, while $\downarrow$ indicates that lower values are preferable.}
\centering
\resizebox{\columnwidth}{!}{%
\begin{tabular}{lccc|ccc}
\toprule
\textbf{Model} & \multicolumn{3}{c|}{\textbf{Simulated Test Data}} & \multicolumn{3}{c}{\textbf{Experimental Data}} \\
 & \textbf{MSE} $\downarrow$ & \textbf{SNR} $\uparrow$ & \textbf{COR} $\uparrow$ & \textbf{MSE} $\downarrow$ & \textbf{SNR} $\uparrow$ & \textbf{COR} $\uparrow$ \\
\midrule
$\beta$-VAE 0.5 & 6.58e-3 & 14.74 & 0.19 & 8.17e-3 & 15.29 & 0.18 \\
$\beta$-VAE 0.7 & 8.17e-3 & 13.89 & 0.18 & 1.19e-2 & 13.77 & 0.16 \\
$\beta$-VAE 1.0 & 9.95e-3 & 12.86 & 0.16 & 1.33e-2 & 13.03 & 0.14 \\
WAE          & \textbf{2.40e-4} & \textbf{28.99} & \textbf{0.67} & \textbf{3.70e-4} & \textbf{28.38} & \textbf{0.78} \\
\bottomrule
\end{tabular}
}

\label{tab:wae_vae_metrics}
\end{table}

Table~\ref{tab:wae_vae_metrics} summarizes the reconstruction performance and latent space structure of different auto-encoding models evaluated on both test set pulses and experimental measurements. Specifically, we compare the obtained WAE with $\beta$-VAEs of the same network layout as shown in Figure \ref{fig:wae}, with the training objective of ELBO maximization, Gaussian prior $\mathcal{N}(0, I)$ and sampling via reparameterization trick in the latent space.
Next to the MSE loss, we further assess reconstruction fidelity using signal-to-noise ratio (SNR) 
\[
\mathrm{SNR} = \frac{1}{N} \sum_{i=1}^{N} 10 \log_{10} \left( \frac{||x_i||^2}{||x_i-\hat{x}_i||^2} \right).
\]
The WAE achieves SNR values of 28–29 dB on both test and experimental pulses, indicating high-fidelity reconstructions that preserve fine-scale structure and coinciding with the visual inspection in Figure \ref{fig:recon}. In contrast, $\beta$-VAEs attain at best 14 dB, demonstrating significant information loss.

To quantify how well the latent representations preserve distances in pulse variations, we compute the Pearson correlation (COR) between pairwise Euclidean distances in data space and distances in latent space. For $N-$batched pulse samples $\{x_i\}_{i=1}^N$ we compute pairwise distances $d_X^{ij} := ||x_i-x_j||$ and $d_Z^{ij} := ||z_i-z_j||$ and average
\[
\frac{\mathrm{Cov}(d_X, d_Z)}{\sigma_{d_X} \sigma_{d_Z}}
\]
over a total of $50$ batches of size $128$. The WAE achieves values of $0.70$ on held-out simulation data and $0.78$ on experimental pulses, indicating strong alignment between latent and data-space geometries. In contrast, $\beta$-VAEs trained with identical architectures yield substantially lower scores in the range $0.16–0.19$, reflecting significant distortion of relative distances due to per-sample latent regularization.

\section{Downstream Beam Dynamics Simulations}

To assess the downstream relevance of the learned latent pulse representations, we performed beam dynamics simulations with the code ASTRA \cite{Astra}. To this end, we linearly interpolated between randomly drawn latent codes and sampled electron emission time histograms from the corresponding decoded pulse shapes as described in section \ref{subsec:linear_transform_sampling} and shown in appendix \ref{appendix:sampling}. Electron beam propagation was simulated along a photoinjector beamline up to approximately $4.5\,\mathrm{m}$, resulting in a total of $5\times10^4$ simulation runs spanning diverse pulse shapes, varying longitudinal bunch lengths and machine parameter settings.

\begin{figure}
    \centering
    \includegraphics[width=\linewidth]{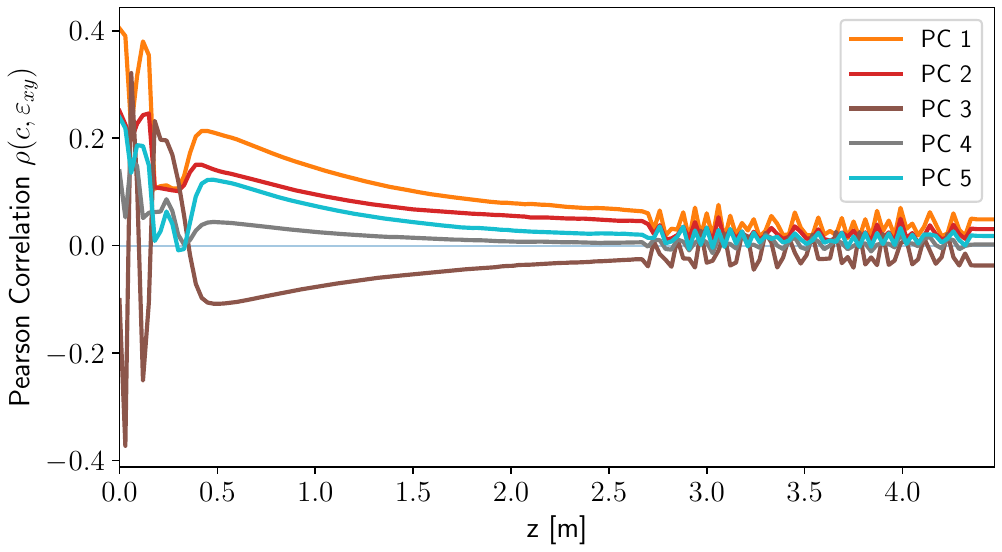}
    \caption{Linear correlations between the first 5 latent principal components and $\varepsilon_{xy}$. Oscillations between $2.5\,\mathrm{m}$ and $4.5\,\mathrm{m}$ arise from an accelerating module within the accelerator lattice, causing fluctuations in transverse momenta.}
    \label{fig:downstream_correlations}
\end{figure}

As a figure of merit, we consider the normalized transverse emittance $\varepsilon_{xy}$ \cite{PhysRevSTAB.6.034202}, which quantifies the transverse phase-space volume of the electron bunch and is a central objective in photoinjector optimization. Linear correlations $\rho(c_i,\varepsilon_{xy}(z))$ along the beamline with the first five principal components $c_1,\dots,c_5$ of the WAE latent space as shown in Figure \ref{fig:downstream_correlations} reveal a significant impact of the pulse shape on the transverse emittance in the near-cathode, space-charge-dominated regime, aligning with physical expectations. Since attributes such as pulse length and amplitude are removed by preprocessing, these dependencies arise from differences in pulse shape solely. Further downstream, correlations decay as accelerator fields increasingly govern beam evolution.

\section{Conclusions}

This work establishes a data-driven framework for learning geometrically faithful and physically meaningful latent representations of longitudinal laser pulse shapes. We demonstrated that Wasserstein-based representation learning yields latent coordinates that preserve meaningful distances between laser pulse shapes and support smooth latent interpolations.

Beyond evaluating reconstruction and latent space quality, we validated the physical relevance of the learned latent space by embedding experimental pulse sequences and by propagating generated pulse shapes through beam dynamics simulations. The resulting coherence of experimental trajectories within the simulated latent manifold, as well as the observed correlations between latent coordinates and downstream emittance metrics, indicate that the representation captures pulse variations that are meaningful for accelerator operation. These results suggest that the latent space could act as a physically structured coordinate system over which interpolation and analysis can be performed efficiently, without requiring repeated evaluation of forward pulse simulations.

While this study focused on longitudinal pulse shaping, the proposed approach is not tied to a specific laser system or accelerator configuration. The representation is learned solely from distributions of physically admissible signals and can therefore be generalized to other pulse shaping modalities, including transverse and spatiotemporal profiles. As such, the latent space can serve as a reusable interface between optical shaping systems and downstream optimization or control tasks.



\newpage

\section*{Impact Statement}

This work contributes a data-driven representation learning framework for modeling and analyzing laser pulse shapes in accelerator-based light sources. By enabling structured exploration of pulse shape variability through a learned latent space, the approach has the potential to support more efficient simulation workflows and reduce simulation overhead in scientific research settings. While the methods are developed and evaluated in a controlled accelerator physics context, they are not intended for direct operational control without further validation; responsible use requires careful integration with domain knowledge and safety constraints. More broadly, the framework illustrates how generative models can serve as interpretable interfaces between complex physical subsystems, a paradigm that may generalize to other scientific and engineering domains.

\section*{Acknowledgements}

This work was supported by the German Federal Ministry of Research, Technology and Space (formerly Federal Ministry of Education and Research) under project OPAL-FEL, grant number 05D23GT1. The authors thank the FS-LA and MXL groups at DESY in Hamburg, Germany. They also acknowledge the support of their project partners at the PITZ facility in Zeuthen, Germany. In addition, the authors acknowledge DESY, a member of the Helmholtz Association (HGF), for institutional support and for providing access to the Maxwell computational resources operated at DESY.

\nocite{*}

\bibliography{sources}
\bibliographystyle{icml2026}

\newpage
\appendix
\onecolumn
\section{Gaussian Mixture Models in WAE Latent Space}\label{appendix:gmm}

To analyze and structure the latent space learned by the WAE, we fit a 
\textit{Gaussian Mixture Model} (GMM) to the encoded latent representations of simulated pulse shapes. 
Compared to sampling from the standard isotropic prior \( \mathcal{N}(0, I) \), the GMM provides 
a data-adaptive prior that concentrates probability mass around well-trained, physically meaningful 
regions of the latent space. This improves the quality and diversity of generated samples and enables 
cluster-conditional generation for targeted pulse synthesis.

\begin{figure}[!h]
    \centering
    \includegraphics[width=\linewidth]{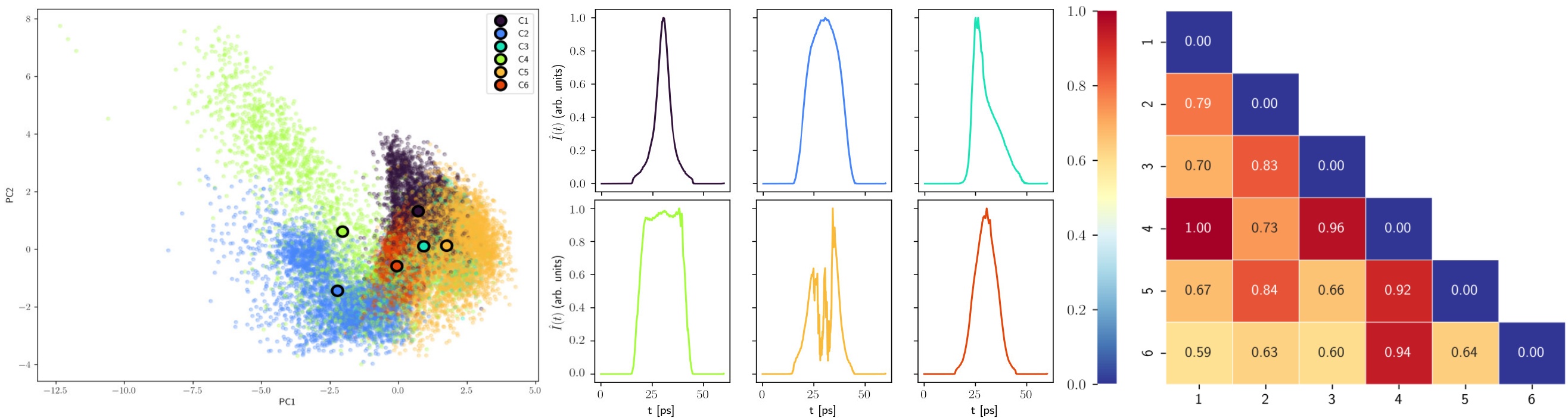}
    \caption{Left panel: PCA projection of the train and test embeddings colored w.r.t. the fit GMM components and their means. 
    Middle panel: Decoded means of the GMM components. Correspondence to components is indicated by the coloring from the left panel. 
    Right panel: Pairwise normalized 2-Wasserstein distances between GMM components.}
    \label{fig:latent_gmm}
\end{figure}

Formally, a GMM models a probability distribution over latent codes \( z \in \mathbb{R}^{d_z} \) as a 
weighted sum of \( K \) multivariate normal components:
\[
p(z) = \sum_{k=1}^K \pi_k \, \mathcal{N}(z \mid \mu_k, \Sigma_k),
\]
where \( \pi_k \) are the mixture weights with \( \sum_{k=1}^K \pi_k = 1 \), and each component is 
parameterized by its mean \( \mu_k \in \mathbb{R}^{d_z} \) and covariance matrix 
\( \Sigma_k \in \mathbb{R}^{d_z \times d_z} \). The parameters are estimated via 
Expectation-Maximization using the latent codes from the whole dataset. The results of a GMM
with $K=6$ components are visualized in Figure \ref{fig:latent_gmm}.

The 2-Wasserstein distance between two multivariate Gaussian components 
\(\mathcal{N}(\mu_i, \Sigma_i)\) and \(\mathcal{N}(\mu_j, \Sigma_j)\), with \(i, j \in \{1, \dots, K\}\), 
can be computed in closed form as
\[
W_{2, ij}^2 = W_2^2\left( \mathcal{N}(\mu_i, \Sigma_i), \mathcal{N}(\mu_j, \Sigma_j) \right) =
\|\mu_i - \mu_j\|^2 + \operatorname{Tr}\left( \Sigma_i + \Sigma_j - 2\left( \Sigma_i^{1/2} \Sigma_j \Sigma_i^{1/2} \right)^{1/2} \right).
\]
This metric allows for quantifying the separation between latent clusters obtained from the 
GMM fit. Moderate to large distances indicate well-separated and 
diverse pulse families in the latent space, while small distances may suggest redundancy 
or overlap between components. The right panel in Figure~\ref{fig:latent_gmm} shows the normalized pairwise Wasserstein $\widetilde{W}_{2, ij}$ distances for the fitted GMM
\[
\widetilde{W}_{2, ij} = \dfrac{W_{2, ij}}{\max\limits_{i,j}\,W_{2, ij}}
\]

As indicated by the measured 2-Wasserstein distances between the components, the most distinct one with the largest
distance to all other components corresponds to flattop shapes, with Gaussian shapes being the closest to these.


\section{Sampling from WAE latent space}
\label{appendix:sampling}

\begin{figure}
    \centering
    \includegraphics[width=\linewidth]{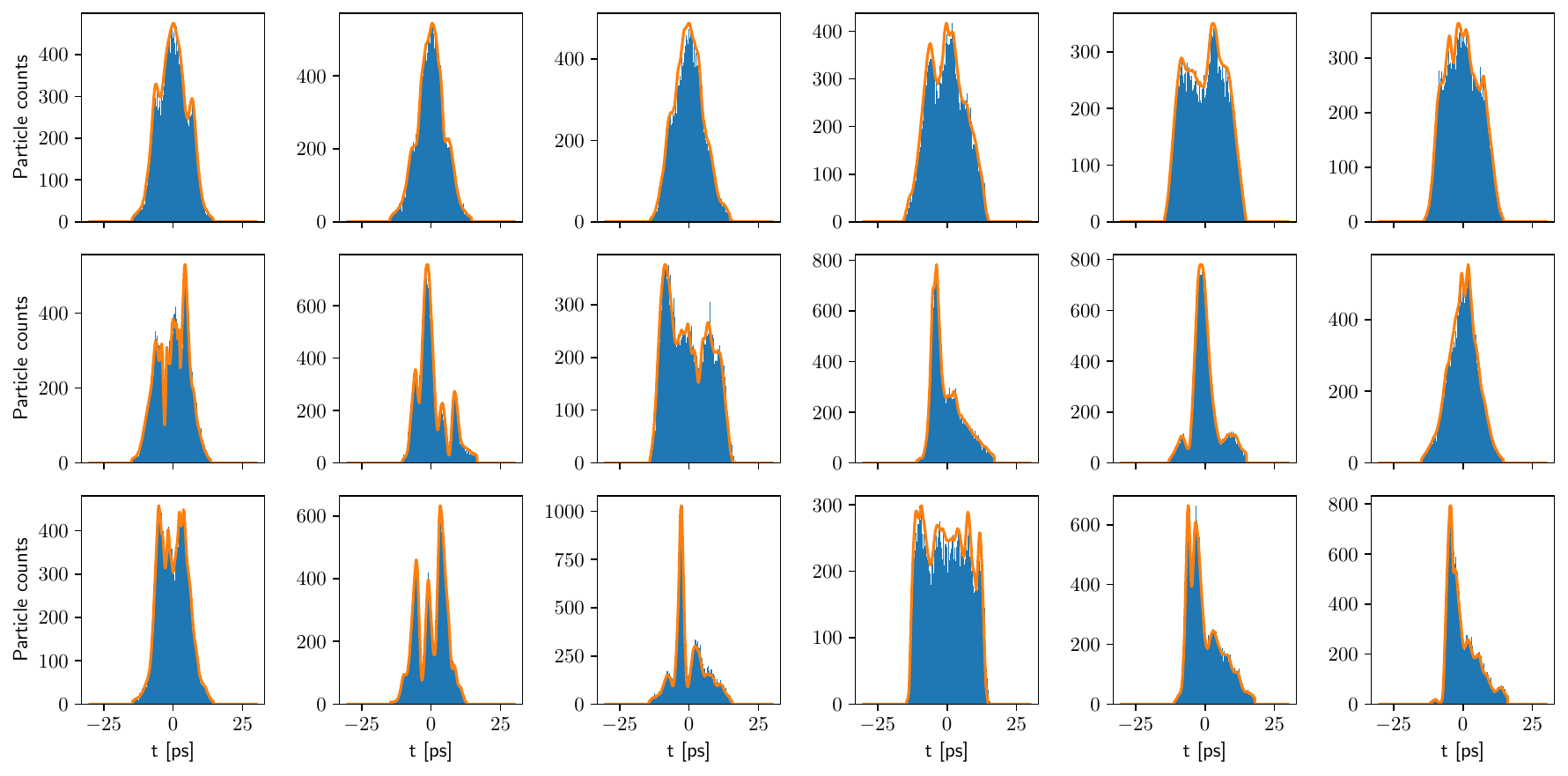}
    \caption{Pulse shapes decoded from latent samples drawn via a latent Gaussian mixture model (orange) are interpreted as probability density functions for electron emission time. Inverse transform sampling produces zero-centered emission time histograms (blue) of $200\,000$ particles each that faithfully match the decoded pulse profiles.}
    \label{fig:latent_its_sampling}
\end{figure}

Beyond interpolation and optimization, the learned latent space enables generative sampling of physically plausible pulse shapes. Latent codes drawn from a Gaussian Mixture Model are decoded into temporal intensity profiles, which are subsequently interpreted as probability density functions for electron emission time. Sampling via inverse transform sampling yields emission time distributions that accurately reproduce the underlying pulse shapes as shown in Figure \ref{fig:latent_its_sampling}. This construction provides a direct and differentiable interface between generative pulse modeling and downstream particle-based beam dynamics simulations such as ASTRA. 

\section{Comparison of WAE with various $\beta$-VAEs}

To evaluate the performance of our proposed WAE, we compared it with different $\beta$-VAEs for $\beta\in\{1.0, 0.7, 0.5\}$. In our experiments the network structure of the VAEs is the same as for the WAE shown in Figure \ref{fig:wae}, only differing
in sampling the latent codes \textit{z} via the reparameterization trick from a learned normal distribution parameterized by the encoder outputs.

\begin{figure}[!h]
    \centering
    \includegraphics[width=\linewidth]{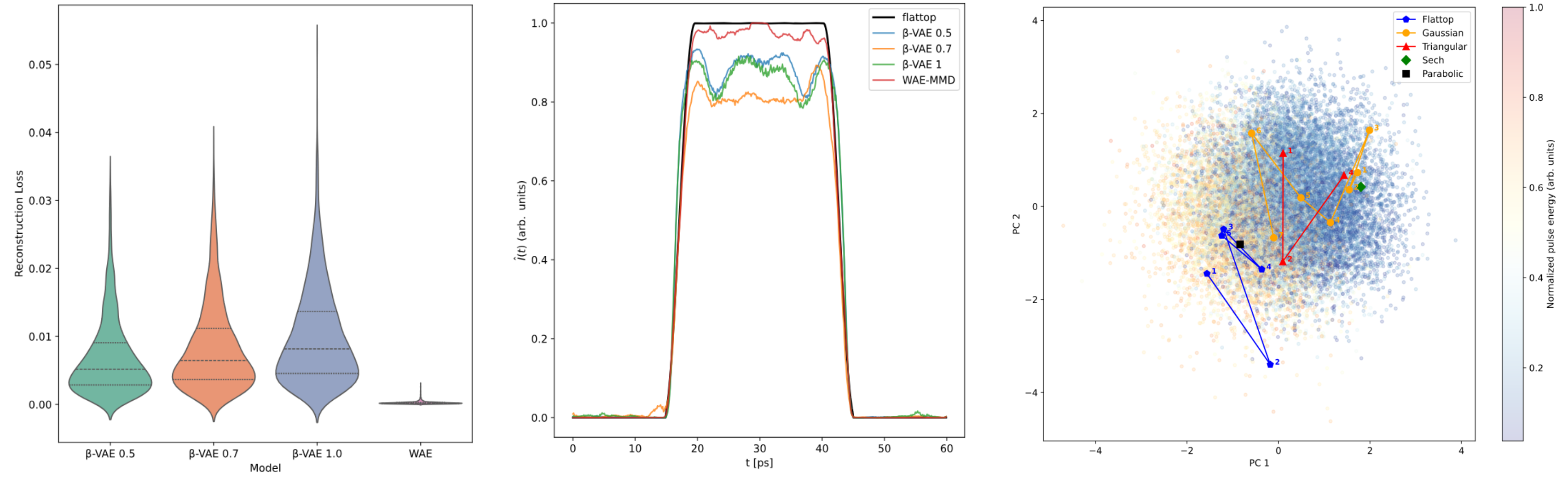}
    \caption{Left panel: Distribution of reconstruction MSE on the test set for different values of $\beta$ and the WAE. Middle panel: Reconstruction of a representative flattop pulse, illustrating progressive loss of sharp temporal features with increasing $\beta$, while the WAE preserves the target shape.
Right panel: PCA projection of the learned latent spaces with trajectories of selected pulse families, showing degraded geometric organization for $\beta$-VAEs compared to the smooth and interpretable structure obtained with the WAE.}
    \label{fig:vae_comparison}
\end{figure}

Figure \ref{fig:vae_comparison} compares the $\beta$-VAEs and the proposed WAE with respect to reconstruction fidelity and latent-space structure. Increasing the $\beta$ coefficient in VAEs leads to systematically higher reconstruction error and visibly degraded reconstructions of sharp pulse features, consistent with excessive regularization and information loss. In contrast, the WAE achieves substantially lower reconstruction error and preserves sharp temporal features critical for beam dynamics. Moreover, visualization of the latent space and pulse family trajectories in a similar fashion as executed for the WAE in Figure \ref{fig:latent_1} reveals that $\beta$-VAEs distort the organization of pulse families, whereas the WAE maintains smooth and interpretable trajectories corresponding to known shape variations. These results indicate that aligning the aggregated posterior rather than enforcing per-sample regularization is crucial for learning geometrically faithful latent representations in this setting. Additional metrics quantifying the reconstruction and latent space quality of the tested models are provided by Table \ref{tab:wae_vae_metrics}.

\end{document}